\documentclass[sigconf]{acmart}

\makeatletter
\@ifundefined{Bbbk}{}{%
  
}
\makeatother

\usepackage{graphicx}
\usepackage{booktabs}
\usepackage{multirow}
\usepackage{tabularx}
\usepackage{array}

\usepackage{amsmath,amssymb}
\usepackage{amsthm}

\usepackage{algorithm}
\usepackage{algorithmic}

\usepackage{xcolor}

\copyrightyear{2026}
\acmYear{2026}

\setcopyright{cc}
\setcctype{by}

\acmConference[MM '26]
{Proceedings of the 34th ACM International Conference on Multimedia}
{November 10--14, 2026}
{Rio de Janeiro, Brazil}

\acmBooktitle{Proceedings of the 34th ACM International Conference on Multimedia
(MM '26), November 10--14, 2026, Rio de Janeiro, Brazil}

\acmISBN{979-8-4007-2213-4/2026/11}
\acmDOI{10.1145/3767308.3834930}

\begin{document}

\title{LoDA: A Level of Detection Aware Method and a Multimodal Sensing Benchmark for Object Level Change Detection}


\author{Haitian Wang}
\affiliation{%
\institution{Western Australia Machine Intelligence Group Pty Ltd}
\city{Nedlands}
\state{WA}
\country{Australia}
}
\affiliation{%
\department{Department of Computer Science and Software Engineering}
\institution{The University of Western Australia}
\city{Crawley}
\state{WA}
\country{Australia}
}
\affiliation{%
\department{School of Electrical Engineering, Computing and Mathematical Sciences}
\institution{Curtin University}
\city{Bentley}
\state{WA}
\country{Australia}
}
\email{haitian.wang@uwa.edu.au}

\author{Xinyu Wang}
\affiliation{%
\institution{Western Australia Machine Intelligence Group Pty Ltd}
\city{Nedlands}
\state{WA}
\country{Australia}
}
\affiliation{%
\department{Department of Computer Science and Software Engineering}
\institution{The University of Western Australia}
\city{Crawley}
\state{WA}
\country{Australia}
}
\email{xinyu.wang@uwa.edu.au}

\author{Sheldon Fung}
\affiliation{%
\department{Department of Computer Science and Software Engineering}
\institution{The University of Western Australia}
\city{Crawley}
\state{WA}
\country{Australia}
}
\email{sheldon.feng@uwa.edu.au}

\author{Xian Zhang}
\affiliation{%
\department{Department of Computer Science and Software Engineering}
\institution{The University of Western Australia}
\city{Crawley}
\state{WA}
\country{Australia}
}
\email{xian.zhang@research.uwa.edu.au}

\author{Zichen Geng}
\authornote{Corresponding author.}
\affiliation{%
\department{Department of Computer Science and Software Engineering}
\institution{The University of Western Australia}
\city{Crawley}
\state{WA}
\country{Australia}
}
\email{zen.geng@research.uwa.edu.au}

\renewcommand{\shortauthors}{Haitian Wang, Xinyu Wang, Sheldon Fung, Xian Zhang, and Zichen Geng}

\begin{abstract}
High-definition 3D LiDAR maps are important for autonomous driving and smart-city services, which require reliable detection of object-level changes in multi-temporal urban LiDAR to keep digital maps aligned with the physical world. Existing approaches from raster height differencing to depth image and point-cloud networks often remain tile-based and threshold-driven, yielding per-point scores without explicit detection limits or consistent object-level labels. We propose an object-level 3D change-detection pipeline that integrates detection-limit-aware registration, geometry-driven object proxies with rule-based semantic and instance segmentation, and displacement cues in height, volume, and surface-normal direction to assign five change labels with confidence. By decoupling registration, geometry, and semantics, the pipeline propagates pose uncertainty into spatially varying detection limits, stabilizes cross-epoch correspondences, and suppresses false changes caused by residual misalignment and density variation. We also present LoDA, a level-of-detection (LoD) aware benchmark for Subiaco district with fused multi-temporal vehicle-LiDAR maps constructed with LiDAR, GNSS, and IMU support, semantic instances, and object-level annotations. On this benchmark, our method achieves 95.0\% accuracy, 90.8\% macro F1, and 83.0\% macro IoU, exceeding the best baseline by 8.7 IoU points and 4.4 F1 points. On the public Urb3DCD-V2 benchmark evaluated under the official point-wise protocol, it reaches 96.81\% mean accuracy and 89.52\% mean change IoU, improving over the strongest reported baselines by 1.36 points in mAcc and 3.18 points in mIoUch.

\end{abstract}

\ccsdesc[500]{Computing methodologies~Scene understanding}

\keywords{Change Detection, Object-level Analysis, Urban 3D Mapping, Level of Detection, Benchmark Dataset}

\maketitle


\begin{figure*}[t]
  \centering
  \includegraphics[width=\textwidth]{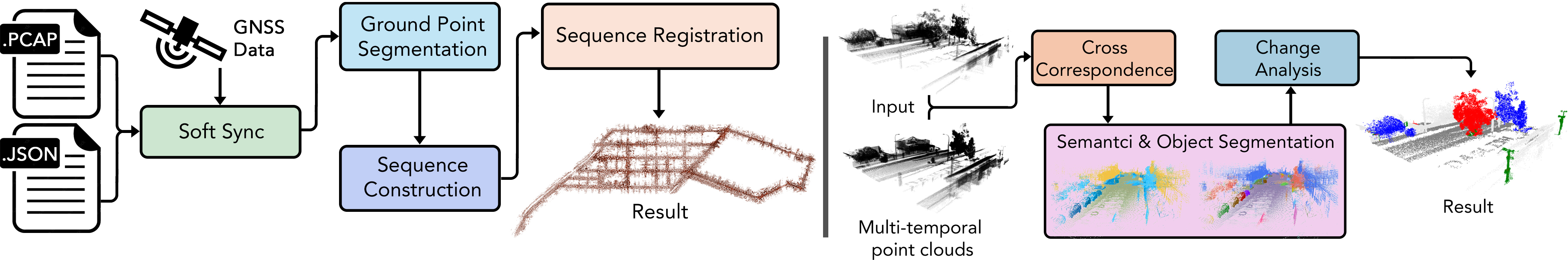}%
  \vspace{-4mm}
  \caption{Overview of the full study. The left part summarises data acquisition and benchmark construction from raw LiDAR packets, metadata, and navigation measurements to fused bi-temporal city maps. The right part shows the proposed change detection method, which takes the fused maps as input, performs LoD-aware correspondence and class-consistent instance formation, and outputs object-level 3D change labels.}
  \Description{A two-part workflow diagram. The upper workflow shows PCAP and JSON files, GNSS data, synchronization, ground segmentation, sequence construction, registration, and a fused urban point-cloud map. The lower workflow shows two temporal point clouds, cross-epoch correspondence, semantic and object segmentation, change analysis, and a color-coded object-level change result.}
  \label{fig:intro-overview}
  \vspace{-5mm}
\end{figure*}

\vspace{-1mm}
\section{Introduction}

High-definition 3D LiDAR maps are a core representation for autonomous driving and smart-city systems because they encode road geometry, traffic context, and static scene structure beyond the current sensor view \cite{seif2018icity,asrat2024hdmaps}. They support map-based localization, long-term perception, and large-scale urban data management \cite{cadena2016slam,caesar2020nuscenes,sun2020waymo,behley2019semantickitti}. In practice, however, urban environments are not static. Buildings are modified, vegetation grows or is pruned, and roadside infrastructure changes over time. These changes must be identified reliably if HD maps are to remain consistent with the physical world.

Existing change detection methods do not fully address this requirement. Classical LiDAR differencing methods compare digital surface models or local point distances, but they are sensitive to registration error and usually report point-wise or surface-wise discrepancies that must be grouped manually \cite{murakami1999alschange,matikainen2010buildingalschange,teo2013lidarchange,lague2013m3c2,xu2021pcdreview}. Raster-based and image-based methods simplify the problem through projection, but they discard vertical structure and rarely model whether a change is actually observable under varying density and range \cite{hussain2013changedetection,daudt2018fcsn}. Recent 3D deep models improve geometric representation learning, yet they typically operate on tiles, rely on fixed thresholds or implicit priors, and output point-wise change predictions rather than structured object-level records \cite{qi2017pointnet,qi2017pointnetplusplus,graham2017submanifold,choy20194d,thomas2019kpconv,degelis2023urb3dcd,degelis2023needschangeinfo}. For map maintenance, the practical requirement is different. The system should decide whether an object has been added, removed, increased, decreased, or left unchanged, and it should do so under an explicit notion of local detectability.

We present an LoD-aware object-level pipeline for bi-temporal vehicle LiDAR maps. It refines cross-epoch alignment and estimates a spatially varying detection limit from local roughness, sampling density, and residual pose uncertainty. Geometry-based object proxies establish stable coarse correspondences before semantic refinement and class-consistent instance formation. LoD-gated height, volume, and normal-direction displacement cues then assign Added, Removed, Increased, Decreased, or Unchanged labels. This design suppresses false changes caused by residual misalignment, density variation, and weak observability while producing structured map-update records.

To evaluate this setting, we introduce LoDA, an object-level LiDAR change-detection benchmark constructed from multimodal vehicle surveys collected in Subiaco in 2023 and 2025 using LiDAR, GNSS, and IMU. LoDA provides fused bi-temporal city maps, semantic and instance annotations, and five-class object-level change labels. Candidate pairs are generated automatically, whereas final labels are independently verified using synchronized annotation views and QA, reducing protocol bias. The benchmark supports controlled object-level evaluation and external transfer testing on Urb3DCD-V2 without per-dataset retuning. Our method achieves $95.0\%$ ACC, $90.8\%$ mF1, and $83.0\%$ mIoU on LoDA, and $96.81\%$ mAcc and $89.52\%$ mIoUch on Urb3DCD-V2. These results demonstrate the value of explicit observability modeling and instance-level reasoning for urban map updating.

Our contributions are:
\begin{itemize}
\item We introduce LoDA, an LoD-aware benchmark for object-level LiDAR change detection, constructed from a multimodal vehicle sensing setup with LiDAR, GNSS, and IMU support, and released with fused bi-temporal city maps, semantic and instance annotations, and five-class object-level change labels.
\item We propose an LoD-aware object-level pipeline that refines cross-epoch alignment, estimates spatially varying detection limits, and builds stable geometry-based correspondences for class-consistent instance reasoning.
\item We introduce LoD-gated object-level change cues that combine height, volume, and normal-direction displacement to infer Added, Removed, Increased, Decreased, and Unchanged for structured map update.
\end{itemize}

\begin{figure*}[t]
  \centering
  \includegraphics[width=\textwidth]{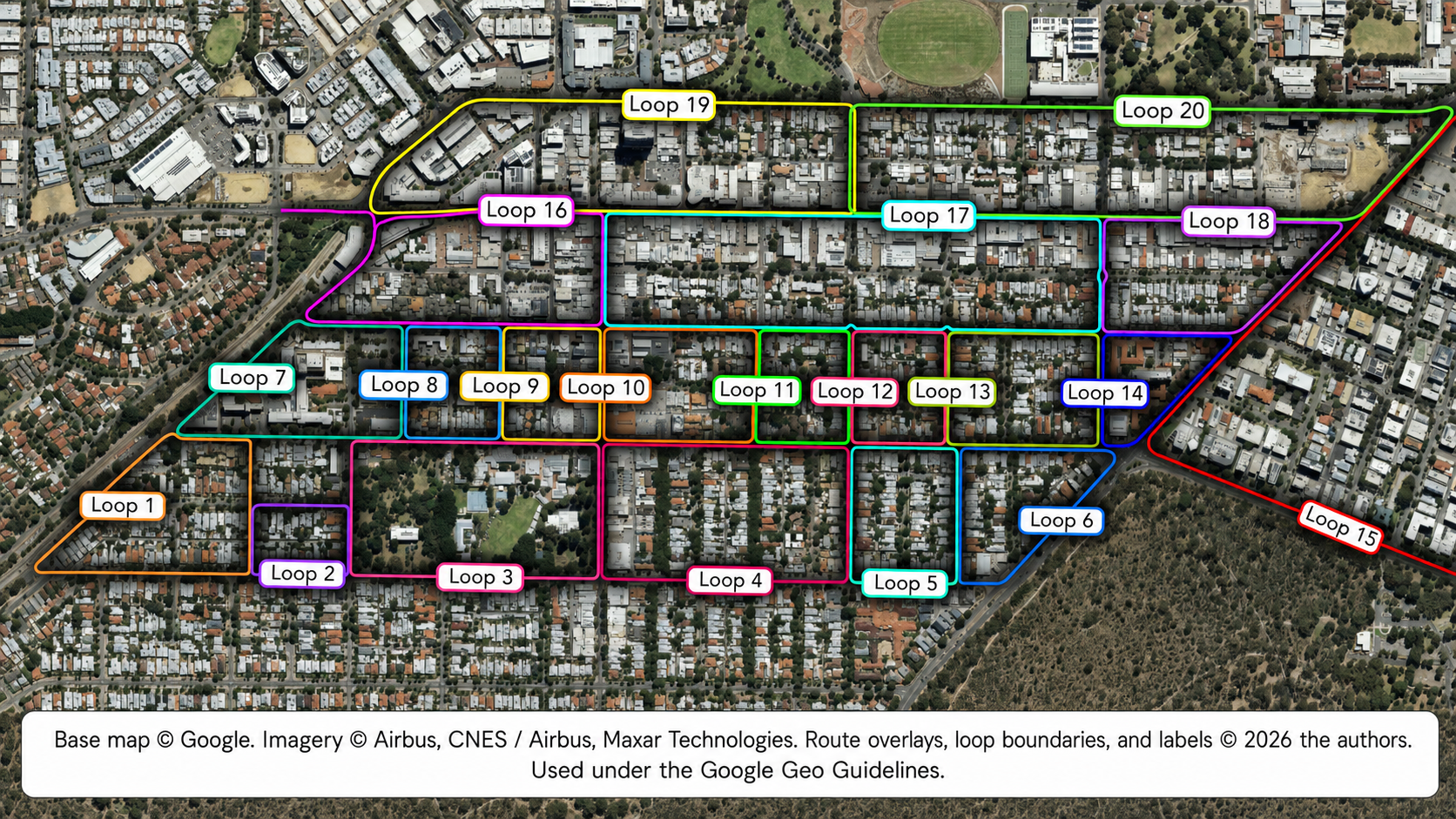}
  \vspace{-4mm}
  \caption{Acquisition layout in Subiaco. The survey is organised into
21 closed loops (Seq 1 to Seq 21) covering main corridors, side streets,
and intersections. Base map \copyright{} Google. Imagery \copyright{}
Airbus, CNES / Airbus, and Maxar Technologies. Route overlays, loop
boundaries, and labels \copyright{} 2026 the authors. Used under the
Google Geo Guidelines.}
  \Description{Satellite map of Subiaco, Western Australia, overlaid with 21 color-coded closed driving loops covering the survey corridors, side streets, and intersections.}
  \vspace{-4mm}
  \label{fig:perth-routes}
\end{figure*}

\vspace{-2mm}
\section{Related Work}\label{sec:related}

Early urban LiDAR mapping largely treated city models as static assets for localization, with updates performed offline or manually \cite{levinson2007maplocalization,seif2018icity}. Classical 3D change detection in laser scanning typically relies on differencing digital surface models or point clouds \cite{teo2013lidarchange}. Raster and height-statistics methods capture large elevation shifts but lose fine 3D structure and are sensitive to occlusion and residual misregistration, while M3C2 measures signed distances along local normals with uncertainty driven by roughness and sampling density but remains a local surface comparator and does not directly yield object-level change records\cite{lague2013m3c2,xu2021pcdreview}. Recent geometric object-level approaches have also started to address long-term LiDAR change reasoning through multi-mission SLAM, probabilistic object differencing, and object correspondence in cluttered environments \cite{rowell2024lista}. Recent variants further reduce uncertainty and improve resolution via patch-based normal-distance estimation \cite{yang2023patchm3c2}.

In remote sensing imagery, change detection is commonly framed as dense 2D mask prediction using spectral differences, object-based classifiers, or Siamese encoder--decoder networks\cite{upreti2022dlcdreview,daudt2018fcsn}. Similar ideas are often transferred to LiDAR via rasterisation into height images or multi-channel grids \cite{teo2013lidarchange}, which simplifies learning but collapses vertical structure and mixes occluded and well-observed regions, typically without modelling spatially varying observability.

Recent 3D deep models learn geometric representations from points or sparse voxels \cite{choy20194d,thomas2019kpconv}. They have been adopted for multi-temporal change detection, including DC3DCD \cite{kharroubi2023dc3dcd} and Urb3DCD with Siamese KPConv-style baselines \cite{degelis2022siamesekpconv,degelis2023urb3dcd}. However, these approaches typically operate on tiles and output per-point change maps, and their robustness can depend on implicit priors learned from labels rather than explicit observability constraints \cite{degelis2023needschangeinfo}.

\vspace{-2mm}
\section{Data Collection}
\label{sec:data-collection}

We collected vehicle LiDAR surveys in Subiaco in 2023 and 2025 to build the bi-temporal city maps used in all experiments. Data were acquired in 21 closed loops with deliberate overlap at intersections, as shown in Fig.~\ref{fig:perth-routes}. The driven distance is $18.6\,\mathrm{km}$ and each loop is logged as an independent sequence at $10$ to $30\,\mathrm{km/h}$. For reproducible evaluation, both epochs are partitioned into fixed $120\,\mathrm{m}\times120\,\mathrm{m}$ blocks aligned to the GDA2020 zone 50 grid. A block is kept if both epochs satisfy at least $40\%$ occupancy on a $0.5$\,m horizontal grid after ground removal and the post-registration block centroid offset is below $0.5$\,m. Block IDs are split into $60\%/10\%/30\%$ train, validation, and test by hashing the block index. LoDA is intended as a controlled urban benchmark for object-level change analysis in vehicle LiDAR rather than an exhaustive cross-city suite, and cross-scene transfer is further examined by applying the same frozen pipeline to Urb3DCD-V2 without per-dataset retuning.

Data were acquired with an Ouster OS1-128 LiDAR, dual-antenna RTK GNSS, and a MEMS IMU under GNSS-PPS synchronization and fixed extrinsic calibration. Packets are decoded into organised scans and deskewed using interpolated IMU poses. Trajectories are estimated on a voxel pyramid $\{1.0,0.5,0.25\}\,\mathrm{m}$ using multi-resolution NDT and robust point-to-plane ICP, with loop closures and GNSS/IMU constraints fused in a pose graph optimised in the GDA2020 zone 50 frame\cite{shan2020liosam,xu2022fastlio2,kissicp2023}. The city map is fused into a global $0.25\,\mathrm{m}$ voxel grid and a $2\,\mathrm{m}$ ground model is estimated for height normalisation. Each epoch provides per-point semantic labels in $\{\text{ground},\text{building},\text{vegetation},\text{mobile}\}$ and per-class instance IDs defined on the fused map. Building instances are obtained by merging adjacent planar parts with compatible facade normals and overlapping footprints. Vegetation and mobile instances follow connectedness under a density-adaptive Euclidean radius.

Let $\mathcal{Y}$ denote the five change labels: \textit{Added}, \textit{Removed}, \textit{Increased}, \textit{Decreased}, and \textit{Unchanged}. The released object labels are produced by an annotation protocol that is independent of the proposed inference rules. Candidate cross-epoch pairs are first proposed automatically under centroid distance below $2$\,m and 3D OBB IoU above $0.05$. Each candidate is then verified during annotation using synchronized plan-view renderings, height maps, and occupancy overlays, and the final label is assigned after QA. Added and Removed are used when an instance is verified to exist in only one epoch. For matched instances, Increased and Decreased are assigned only when visible object-level growth or shrinkage is supported by the annotation views, otherwise Unchanged is used. The local $\mathrm{LoD}_{95}$ map is available during annotation only as an observability aid and does not directly determine the released label. Cases that remain ambiguous after QA are excluded from the benchmark statistics rather than forced to follow the output of the proposed method.

\begin{figure*}[t]
  \centering
  \includegraphics[width=\textwidth]{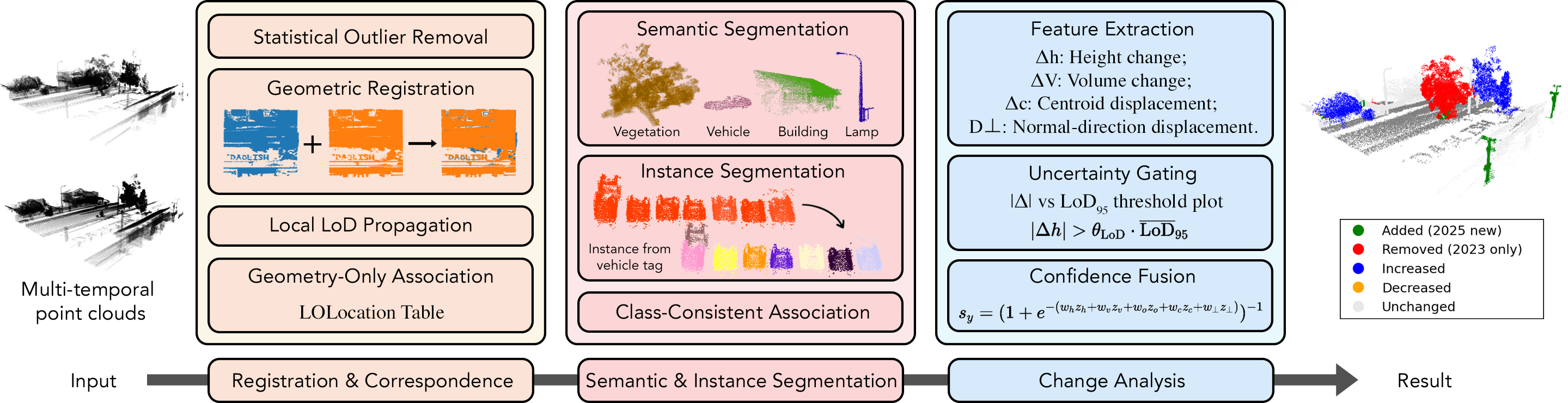}
  \vspace{-6mm}
  \caption{LoD-aware object-level change-detection pipeline. Registration and correspondence produce LoD-aware matches, followed by class-consistent instance formation and LoD-gated five-class change inference.}
  \Description{Pipeline from two temporal point clouds through statistical outlier removal, geometric registration, local level-of-detection propagation, geometry-only association, semantic and instance segmentation, class-consistent association, feature extraction, uncertainty gating, confidence fusion, and five-class object-level change output.}
  \label{fig:pipeline}
  \vspace{-4mm}
\end{figure*}

\vspace{-2mm}
\section{Method}\label{sec:method}

Our method takes a pair of fused bi-temporal city maps as input and performs object-level change inference in three stages. We first refine cross-epoch geometric alignment and estimate a spatially varying level of detection to define local observability. We then construct geometry-based object correspondences and class-consistent instances. Finally, we assign change labels from LoD-gated geometric cues, including height, volume, and normal-direction displacement.

\vspace{-2mm}
\subsection{Geometric Registration and Object-Level Correspondence}
\label{sec:data-layer}

Let $\mathcal{P}^{23}$ and $\mathcal{P}^{25}$ denote the fused Subiaco point clouds from 2023 and 2025. We estimate a rigid alignment $T_{23\rightarrow25}\in\mathrm{SE}(3)$ and a spatially varying normal-direction level of detection (LoD), then build geometry-only object proxies and a cross-epoch correspondence table that is reused by later semantic and instance processing.

Both epochs are filtered by Statistical Outlier Removal ($k=20$, standard-deviation multiplier $1.0$) and downsampled into a voxel pyramid with leaf sizes $\{1.0,0.5,0.25\}$\,m. All thresholds are selected once on the LoDA validation split and then frozen for the LoDA test split and Urb3DCD-V2. Normals are estimated within a radius of $0.75$\,m using at most $30$ neighbours, and near-isotropic neighbourhoods are rejected. Varying $k$ within $[16,24]$ changes validation mIoU by less than $0.3$ points. Pose initialization uses multi-resolution NDT~\cite{biber2003ndt,magnusson2007ndt}, followed by robust point-to-plane ICP~\cite{besl1992icp,chen1992range} over the three pyramid levels. ICP uses correspondence radii $\{1.5,0.75,0.35\}$\,m and at most $\{60,40,30\}$ iterations, with early stopping at pose updates below $10^{-4}$. KD-tree correspondences~\cite{bentley1975kdtree} are retained only when the paired normals differ by less than $30^\circ$.

Using correspondences $\mathcal{C}=\{(p_i,q_i,n_i)\}$ with $p_i\in\mathcal{P}^{23}$, $q_i\in\mathcal{P}^{25}$ and $n_i$ the unit normal at $q_i$, we minimize
\begin{equation}
\label{eq:pt2plane}
\min_{R\in\mathrm{SO}(3),~t\in\mathbb{R}^3}\;
\frac{1}{|\mathcal{C}|}\sum_{(p_i,q_i,n_i)\in\mathcal{C}}
\rho_{\tau}\!\left(\frac{\bigl(n_i^{\top}(Rp_i+t-q_i)\bigr)^2}{\sigma_i^2}\right),
\end{equation}
where $\rho_{\tau}$ is the Tukey biweight~\cite{holland1977robust} with $\tau=0.30$\,m and $\sigma_i^2=\sigma_0^2+\sigma_r^2(q_i)$ combines a fixed sensor term $\sigma_0=0.02$\,m with local roughness $\sigma_r(q_i)$. To compute $\sigma_r(q_i)$, we fit a local plane by PCA in the same $r_n$ neighborhood and collect signed point-to-plane residuals $\{r_\ell\}$. We estimate the robust standard deviation as $\hat{\sigma}_r=1.4826\cdot \mathrm{median}_\ell|r_\ell-\mathrm{median}(r_\ell)|$~\cite{rousseeuw1993mad} and set $\sigma_r^2=\hat{\sigma}_r^2$. Optimization uses Gauss--Newton with rejection of pairs whose unsigned residual exceeds $\tau$. The pose covariance $\Sigma_T\in\mathbb{R}^{6\times6}$ is approximated by the inverse normal-equation matrix at the solution~\cite{censi2007icpcov}, and $\Sigma_t\in\mathbb{R}^{3\times3}$ is taken as the translational block of $\Sigma_T$. We use this translational block in Eq.~\eqref{eq:lod} and neglect the rotational term because evaluation is performed on fixed $120\,\mathrm{m}$ blocks after rigid refinement and vertical normalization, where the residual rotational contribution was empirically smaller than the local roughness term.

To enforce a common vertical axis, we fit ground planes on a $2$\,m grid in each epoch by RANSAC~\cite{fischler1981ransac} with an inlier threshold of $0.10$\,m and a minimum of $200$ inliers per cell. We take the median ground normal as the global up direction, rotate both maps to align it with $\mathbf{e}_z$, then height-normalize by subtracting the local median ground elevation.

LoD is computed on a uniform $1$\,m grid and linearly interpolated to the voxel centers used by later stages. For each grid cell $g$, we pool points within a radius $r_{\mathrm{LoD}}=1.0$\,m around the cell center, estimate the cell normal $n_g$ by PCA, and compute normal-direction residuals to the best-fit plane as in the registration step. We require $N_{23}(g)\ge 30$ and $N_{25}(g)\ge 30$ to form a valid estimate. Otherwise we mark the cell as low-observability and set $\mathrm{LoD}_{95}(g)=0.50$\,m, which prevents downstream stages from asserting changes in undersampled regions. With $\sigma_{23}^2(g)$ and $\sigma_{25}^2(g)$ computed via the MAD-to-std conversion and squared, the $95\%$ LoD along $n_g$ is
\begin{equation}
\label{eq:lod}
\mathrm{LoD}_{95}(g)=1.96\sqrt{\frac{\sigma_{23}^2(g)}{N_{23}(g)}+\frac{\sigma_{25}^2(g)}{N_{25}(g)}+n_g^{\top}\Sigma_t n_g},
\end{equation}
which defines the minimum detectable displacement under local sampling and alignment uncertainty\cite{lague2013m3c2,m3c2ep2021}.

Geometry-only object proxies are built after removing ground points using the $2$\,m ground model. Non-ground points are voxelized at $0.5$\,m and converted into a binary occupancy grid. To close small gaps on thin structures, we apply a 3D morphological closing~\cite{soille2003morphology} consisting of one dilation and one erosion using a $3\times3\times3$ structuring element in voxel space. Connected components are then extracted with 26-connectivity and treated as proxies. For each proxy $o$, we store centroid $c(o)$, an oriented bounding box $B(o)$ from PCA, $h_{95}(o)$ as the $95$th percentile of height above the local ground surface, and a normalized $10$-bin height histogram $H(o)$ on $[0,h_{95}(o)]$ with uniform bin edges. Proxies with fewer than $200$ occupied voxels are discarded.

Cross-epoch proxy association is solved by gated bipartite assignment. Candidate pairs must satisfy $\lVert c_i-c_j\rVert_2<2.0$\,m and $\mathrm{IoU}(B_i,B_j)>0.10$. Each candidate is scored by
$\mathsf{cost}(o_i,o_j)=w_p \lVert c_i-c_j\rVert_2^2+w_b\bigl(1-\mathrm{IoU}(B_i,B_j)\bigr)+w_h D_{\chi^2}(H_i,H_j)$
with $(w_p,w_b,w_h)=(1.0,2.0,0.5)$. The final assignment uses the Hungarian algorithm~\cite{munkres1957assignment} with a fixed unmatched cost of $3.0$. Matched pairs yield a correspondence table storing $(c,B,h_{95},H)$ for both epochs and the median $\mathrm{LoD}_{95}$ over voxel centers in $B_i\cup B_j$, while unmatched proxies are retained as Added or Removed candidates.

\vspace{-2mm}
\subsection{Semantic and Instance Segmentation}
\label{sec:semantic-layer}

\begin{figure}[t]
  \centering
  \includegraphics[width=\columnwidth]{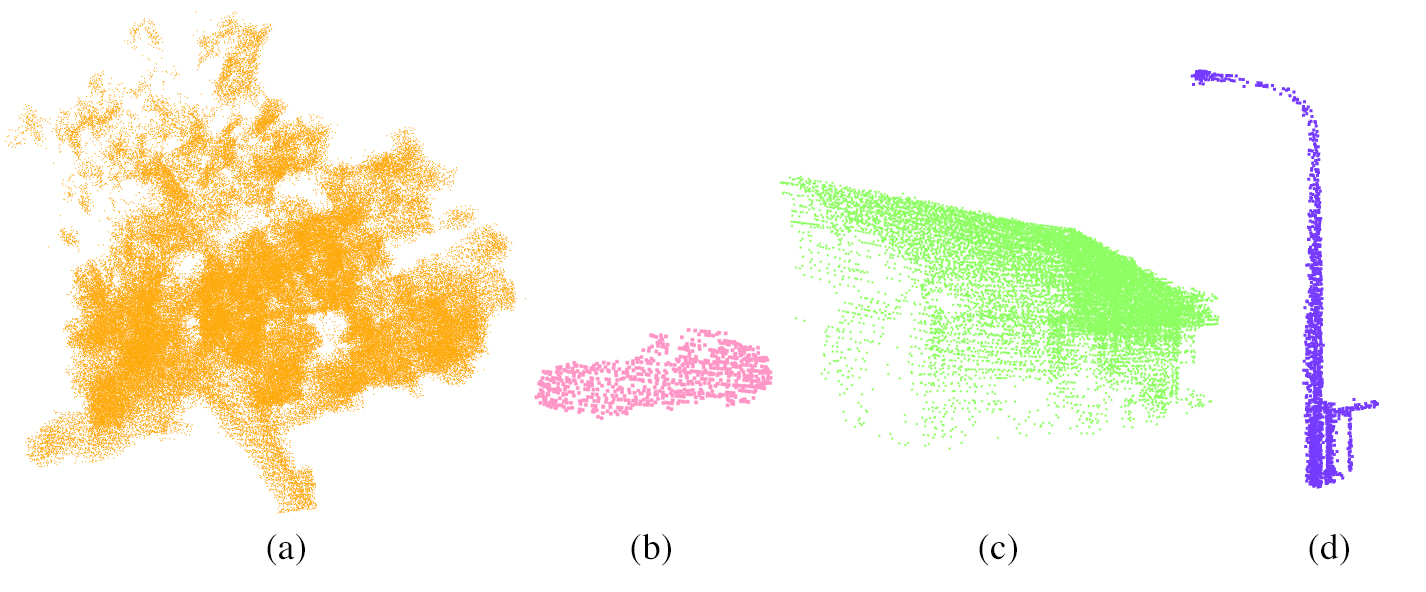}
  \vspace{-4mm}
  \caption{Representative geometry patterns used in the semantic stage: (a) vegetation, (b) compact object, (c) facade-like planar structure, and (d) slender vertical structure.}
  \Description{Four point-cloud examples illustrating vegetation, a compact vehicle-like object, a planar building facade, and a slender vertical pole-like structure.}
  \label{fig:semantic-types}
  \vspace{-4mm}
\end{figure}

We segment each epoch into semantic classes and object instances using geometry cues computed on the $0.25$\,m voxel grid\cite{landrieu2018spg}. For every point, we estimate the local covariance in a $0.6$\,m radius neighborhood and obtain eigenvalues $(\lambda_1 \ge \lambda_2 \ge \lambda_3)$ to derive linearity $L=(\lambda_1-\lambda_2)/\lambda_1$, planarity $P=(\lambda_2-\lambda_3)/\lambda_1$, and sphericity $S=\lambda_3/\lambda_1$. Together with the absolute vertical normal component and height above the ground surface, each point is represented by
\begin{equation}
\mathbf{f}_i=\bigl[L_i,\,P_i,\,S_i,\,|\mathbf{n}_i\!\cdot\!\mathbf{e}_z|,\,z_{g,i}\bigr]^{\top}.
\end{equation}
We then form superpoints by cut-pursuit on a $k$-NN graph with $k=20$ by solving\cite{landrieu2017cutpursuit}. We build the $k$-NN graph on voxel centers at $0.25$\,m with $k=20$ and an edge-length cap of $1.2$\,m to avoid long-range links across streets. Features are standardized per epoch to zero mean and unit variance before graph construction, and we set $\beta=0.8$ and $\sigma_f$ to the median $\|\mathbf{f}_i-\mathbf{f}_j\|_2$ over graph edges.
\begin{equation}
\label{eq:cp}
\min_{\mathcal{U}}
\sum_{u\in\mathcal{U}}\sum_{i\in u}\bigl\|\mathbf{f}_i-\boldsymbol{\mu}_u\bigr\|_1
+\beta\!\!\sum_{(i,j)\in\mathcal{E}}\!\!w_{ij}\,
\mathbb{1}\!\bigl[\mathcal{U}(i)\neq\mathcal{U}(j)\bigr],
\end{equation}
Cut-pursuit iterations stop when the relative decrease in the objective falls below $10^{-4}$ or after $20$ refinement rounds, which produced superpoints of median size 350 points on LoDA, where $\boldsymbol{\mu}_u$ is the segment median feature and $w_{ij}=\exp\!\left(-\|\mathbf{f}_i-\mathbf{f}_j\|_2^2/\sigma_f^2\right)$. Semantic labels are assigned per superpoint and propagated to points with four classes. A superpoint is ground if its median distance to the ground model is below $0.10$\,m and at least $80\%$ of its normals satisfy $|\mathbf{n}\cdot\mathbf{e}_z|>0.996$ (within $5^\circ$ of horizontal). A non-ground superpoint is labeled as building if $P>0.60$, $|\mathbf{n}\cdot\mathbf{e}_z|<0.25$, the plane-fit RMSE is below $0.08$\,m, and the estimated planar support area exceeds $5$\,m$^2$ after merging adjacent coplanar superpoints with normal deviation below $12^\circ$. It is labeled as vegetation if $S>0.35$, median height above ground satisfies $z_g>0.5$\,m, and the within-segment normal azimuth spread between the $95$th and $5$th percentiles exceeds $30^\circ$. Remaining non-ground superpoints are labeled as mobile if their oriented bounding box satisfies length $<5$\,m, width $<3$\,m, height $<3$\,m, and voxelized volume below $60$\,m$^3$. If multiple rules are satisfied, the priority order is ground, building, vegetation, mobile. These class rules were fixed on the LoDA validation split after ground normalization and then used unchanged on the LoDA test split and Urb3DCD-V2, with no per-block retuning. Representative geometry patterns used by the semantic stage are shown in Fig.~\ref{fig:semantic-types}.

Instances are extracted within each class. Buildings are obtained by first extracting planar patches by region growing on superpoints with $P>0.60$ and merging adjacent patches when their normals differ by at most $12^\circ$ and their projected footprints have IoU above $0.30$. Patches with fewer than $800$ points are absorbed into the nearest compatible building instance if the centroid-to-plane distance is below $0.20$\,m, otherwise they are discarded as fragments. Vegetation instances are produced by Euclidean clustering on non-ground, non-building points with an adaptive radius $r_c=\max(0.8\,\mathrm{m},\,1.8/\sqrt[3]{\rho})$, where $\rho$ is the local point density estimated in a $1$\,m ball, and with a minimum of $200$ points per cluster. Mobile instances are produced by DBSCAN~\cite{ester1996dbscan} in the horizontal plane with $\varepsilon=0.7$\,m and $\mathrm{minPts}=30$, followed by eroding the ground mask by $0.15$\,m in 2D to remove clusters connected to curbs and facades. Learning-based moving-object segmentation from sequential LiDAR scans provides an alternative for scan-level dynamic filtering \cite{chen2021moving}, whereas our benchmark operates on fused bi-temporal maps and therefore uses deterministic geometry-based rules for mobile instance extraction.


\vspace{-2mm}
\subsection{Change Analysis}
\label{sec:change-analysis}

\begin{table*}[t]
\centering
\caption{Comparison on LoDA. ACC is micro accuracy; mF1 and mIoU are five-class macro scores, followed by per-class IoU (\%).}
\vspace{-4mm}
\label{tab:method_comp_perth}
\footnotesize
\setlength{\tabcolsep}{3.2pt}
\renewcommand{\arraystretch}{1.05}

\begin{tabular*}{\textwidth}{@{\extracolsep{\fill}}%
>{\raggedright\arraybackslash}p{0.12\textwidth}
>{\raggedright\arraybackslash}p{0.24\textwidth}
cccccccc@{}}
\toprule
Representation & Method & ACC & mF1 & mIoU & Added & Removed & Increased & Decreased & Unchanged \\
\midrule
\multirow{5}{*}{Raster DSM}
& DSM RF~\cite{breiman2001random}                   & 90.6 & 71.0 & 53.2 & 69.8 & 53.4 & 24.8 & 29.7 & 88.2 \\
& DSM Siamese~\cite{daudt2018fcsn}                  & 91.2 & 72.4 & 55.5 & 72.3 & 56.2 & 27.9 & 31.5 & 89.7 \\
& DSM FC EF~\cite{daudt2018fcsn}                    & 91.5 & 72.0 & 55.2 & 73.1 & 55.0 & 26.8 & 30.6 & 90.4 \\
& DSM FC Siam Conc~\cite{daudt2018fcsn}             & 91.9 & 72.9 & 56.6 & 74.5 & 56.3 & 28.7 & 32.4 & 90.9 \\
& DSM U Net~\cite{cicek20163dunet}                  & 92.4 & 74.1 & 58.1 & 76.2 & 58.1 & 30.6 & 34.5 & 91.3 \\
\midrule
\multirow{9}{*}{Point based}
& M3C2~\cite{lague2013m3c2}                         & 92.7 & 75.2 & 59.8 & 78.4 & 60.2 & 33.5 & 36.1 & 90.8 \\
& PointNet++ Siam~\cite{qi2017pointnetplusplus}     & 93.4 & 80.1 & 66.2 & 82.7 & 68.4 & 41.8 & 46.0 & 92.1 \\
& SiamKPConv~\cite{thomas2019kpconv}                & 94.0 & 84.5 & 71.1 & 85.9 & 73.2 & 49.3 & 53.7 & 93.5 \\
& TripletKPConv~\cite{degelis2023urb3dcd}           & 94.3 & 85.7 & 73.0 & 87.4 & 75.1 & 52.6 & 56.9 & 93.0 \\
& EFS KPConv~\cite{kharroubi2023dc3dcd}             & 94.6 & 86.4 & 74.3 & 88.3 & 76.8 & 54.1 & 58.4 & 93.8 \\
& Siam PTv3~\cite{wu2023ptv3}                       & 94.8 & 88.0 & 76.9 & 88.2 & 77.8 & 62.0 & 62.7 & 94.0 \\
& Siam PointMamba~\cite{ma2024pointmamba}           & 94.9 & 88.5 & 77.6 & 89.0 & 78.4 & 63.0 & 63.5 & 94.1 \\
& Siam Mamba3D~\cite{zhang2024mamba3d}               & 94.8 & 88.2 & 77.1 & 88.6 & 78.1 & 62.6 & 62.2 & 94.0 \\
& Siam LitePT~\cite{huang2025litept}                & 94.7 & 87.5 & 76.2 & 88.0 & 77.2 & 60.5 & 61.3 & 94.0 \\
\midrule
\multirow{6}{*}{Voxel based}
& Urb3DCD OneConvFusion~\cite{degelis2023urb3dcd}   & 93.7 & 82.3 & 68.9 & 84.1 & 70.5 & 46.7 & 50.2 & 93.2 \\
& Siam MinkUNet~\cite{choy20194d}            & 93.9 & 84.1 & 70.6 & 85.2 & 72.9 & 48.9 & 52.8 & 93.4 \\
& EFSKPConv~\cite{kharroubi2023dc3dcd}              & 94.1 & 85.2 & 72.3 & 86.7 & 74.4 & 51.3 & 55.6 & 93.6 \\
& MinkUNeXt~\cite{cabrera2025minkunext}               & 93.9 & 84.1 & 70.6 & 85.2 & 72.9 & 48.9 & 52.8 & 93.4 \\
& Siam VoxelMamba~\cite{zhang2024voxelmamba}        & 94.8 & 88.4 & 77.4 & 88.9 & 78.6 & 62.9 & 61.8 & 94.8 \\
& Siam UniMamba~\cite{li2025unimamba}               & 94.9 & 88.8 & 77.9 & 89.1 & 79.0 & 63.8 & 63.0 & 94.6 \\
\midrule
Object based
& \textbf{Ours}                                     & \textbf{95.0} & \textbf{90.8} & \textbf{83.0} & \textbf{87.8} & \textbf{81.2} & \textbf{77.2} & \textbf{73.7} & \textbf{95.1} \\
\bottomrule
\end{tabular*}
\vspace{-4mm}
\end{table*}

\begin{table}[t]
\centering
\caption{Performance on 15 representative LoDA blocks. The final row aggregates all evaluated instances.}
\vspace{-4mm}
\label{tab:perth_blocks}
\scriptsize
\setlength{\tabcolsep}{4.0pt}
\renewcommand{\arraystretch}{1.05}
\resizebox{\columnwidth}{!}{%
\begin{tabular}{lcccccccc}
\toprule
\multirow{2}{*}{Block} & \multirow{2}{*}{ACC} & \multirow{2}{*}{mF1} & \multirow{2}{*}{mIoU} & \multicolumn{5}{c}{Per class IoU (\%)} \\
\cmidrule(lr){5-9}
 &  &  &  & Added & Removed & Increased & Decreased & Unchanged \\
\midrule
C01 & 95.4 & 91.2 & 83.6 & 88.9 & 80.4 & 76.4 & 74.3 & 98.0 \\
C02 & 95.7 & 90.5 & 82.9 & 87.4 & 82.2 & 77.3 & 71.6 & 96.0 \\
C03 & 94.6 & 90.1 & 82.2 & 88.5 & 80.1 & 76.3 & 72.5 & 93.7 \\
C04 & 95.9 & 91.6 & 84.0 & 89.6 & 82.9 & 77.0 & 74.8 & 95.5 \\
C05 & 95.1 & 90.3 & 82.8 & 88.7 & 81.6 & 75.8 & 73.2 & 94.6 \\
C06 & 94.8 & 89.7 & 82.0 & 86.5 & 81.9 & 77.5 & 71.1 & 92.8 \\
C07 & 95.3 & 90.4 & 83.2 & 88.0 & 81.3 & 76.9 & 73.8 & 95.8 \\
C08 & 94.7 & 89.8 & 82.1 & 86.9 & 80.5 & 77.2 & 71.7 & 93.9 \\
C09 & 95.6 & 91.0 & 83.4 & 88.4 & 81.0 & 78.0 & 73.9 & 95.7 \\
C10 & 94.9 & 89.6 & 82.3 & 87.3 & 80.2 & 76.1 & 71.4 & 96.3 \\
C11 & 95.8 & 91.4 & 83.8 & 88.8 & 82.7 & 78.3 & 74.6 & 95.0 \\
C12 & 95.2 & 90.9 & 83.1 & 87.9 & 81.8 & 77.4 & 73.5 & 94.9 \\
C13 & 94.5 & 89.4 & 81.8 & 87.1 & 80.7 & 75.9 & 70.8 & 94.4 \\
C14 & 95.5 & 91.1 & 83.5 & 89.2 & 81.5 & 77.8 & 73.6 & 95.3 \\
C15 & 94.4 & 89.2 & 81.7 & 86.8 & 80.9 & 76.2 & 70.9 & 94.0 \\
\midrule
\textbf{all} & \textbf{95.0} & \textbf{90.8} & \textbf{83.0} & \textbf{87.8} & \textbf{81.2} & \textbf{77.2} & \textbf{73.7} & \textbf{95.1} \\
\bottomrule
\end{tabular}
}
\vspace{-4mm}
\end{table}

Let $\mathcal{C}=\{\text{ground},\text{building},\text{vegetation},\text{mobile}\}$.
For each semantic class $c\in\mathcal{C}$ and each associated pair $(o_i^{23},o_j^{25})$, we evaluate geometry on a voxel grid with edge length $0.5$\,m inside the union of their oriented bounding boxes. For every occupied grid point $x$ we query the nearest surface points $p_{23}(x)$ and $p_{25}(x)$ and their normals $n_{23}(x)$ and $n_{25}(x)$. These samples define the occupied volumes $V^{23}$ and $V^{25}$, the 3D occupancy intersection over union $\mathrm{IoU}_{3\mathrm{D}}$, the centroid shift $\Delta\mathbf{c}=\mathbf{c}_j-\mathbf{c}_i$, the height difference $\Delta h = h_{95}^{25}-h_{95}^{23}$ based on the $95$th percentile height, and the volume change $\Delta V = V^{25}-V^{23}$. Normal direction displacement is estimated as
\begin{equation}
D_{\perp}(o_i,o_j)
=
\operatorname{median}_{x\in\Omega(o_i,o_j)}
\bar{n}(x)^{\top}\bigl(p_{25}(x)-p_{23}(x)\bigr)
\label{eq:normdisp}
\end{equation}
with $\Omega(o_i,o_j)$ the set of voxel centers in the box intersection and $\bar{n}(x)=\bigl(n_{23}(x)+n_{25}(x)\bigr)/\bigl\|n_{23}(x)+n_{25}(x)\bigr\|_2$. The pairwise detection limit in normal direction is
\begin{equation}
\overline{\mathrm{LoD}}_{95}(o_i,o_j)
=
\operatorname{median}_{x\in\Omega(o_i,o_j)}
\mathrm{LoD}_{95}(x)
\end{equation}
using the grid based $\mathrm{LoD}_{95}$ from Eq.~\eqref{eq:lod}. Statistics $\Delta h$ and $D_{\perp}$ are used only if $|\Delta h|>\alpha_{\mathrm{LoD}}\overline{\mathrm{LoD}}_{95}$ and $|D_{\perp}|>\alpha_{\mathrm{LoD}}\overline{\mathrm{LoD}}_{95}$ with $\alpha_{\mathrm{LoD}}=1.10$, selected by a small sensitivity sweep over $\{1.0,1.1,1.2\}$ on the validation split, where performance was stable and $\alpha_{\mathrm{LoD}}=1.10$ avoided spurious \textit{Increased}/\textit{Decreased} flips. Nearest-neighbor queries for $p_{23}(x)$ and $p_{25}(x)$ use a radius cap of $0.6$\,m. If fewer than $50$ valid samples exist in $\Omega(o_i,o_j)$ after radius capping, we set $D_{\perp}(o_i,o_j)=0$ and mark the cue as non-informative so that the decision falls back to IoU, centroid shift, and voxel volume.

\begin{figure}[t]
  \includegraphics[width=0.48\textwidth,height=0.12\textheight,keepaspectratio=false]{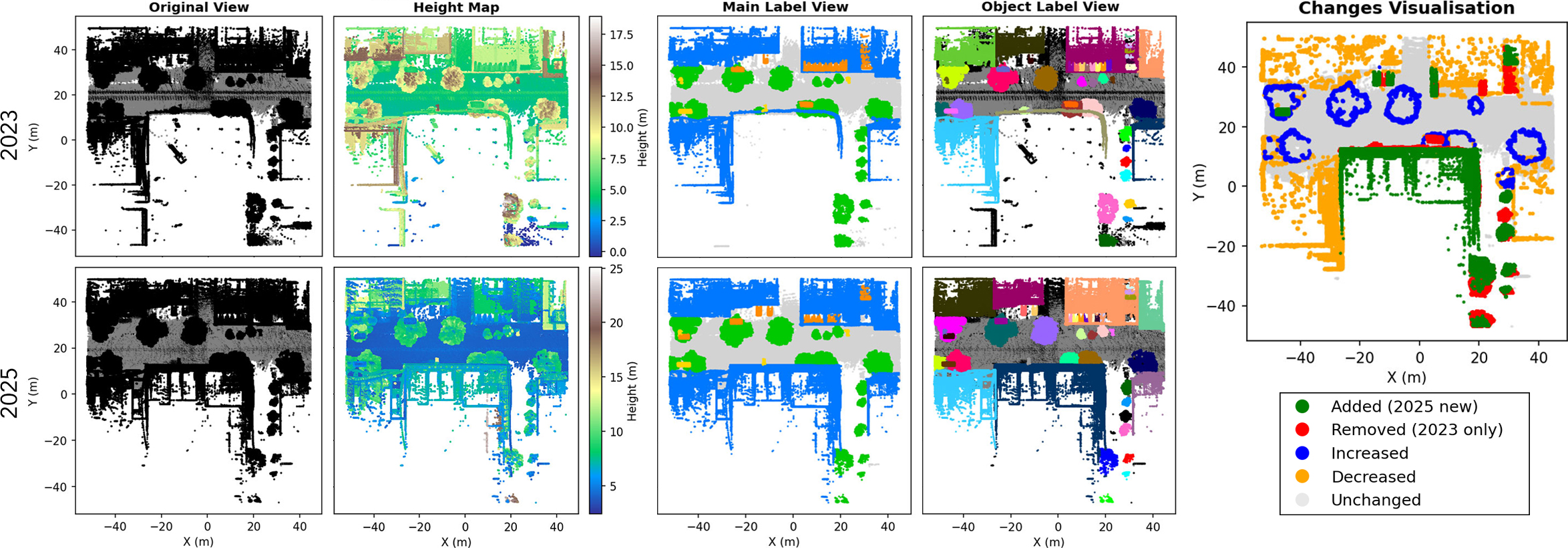}\\[-1pt]
  \includegraphics[width=0.48\textwidth,height=0.12\textheight,keepaspectratio=false]{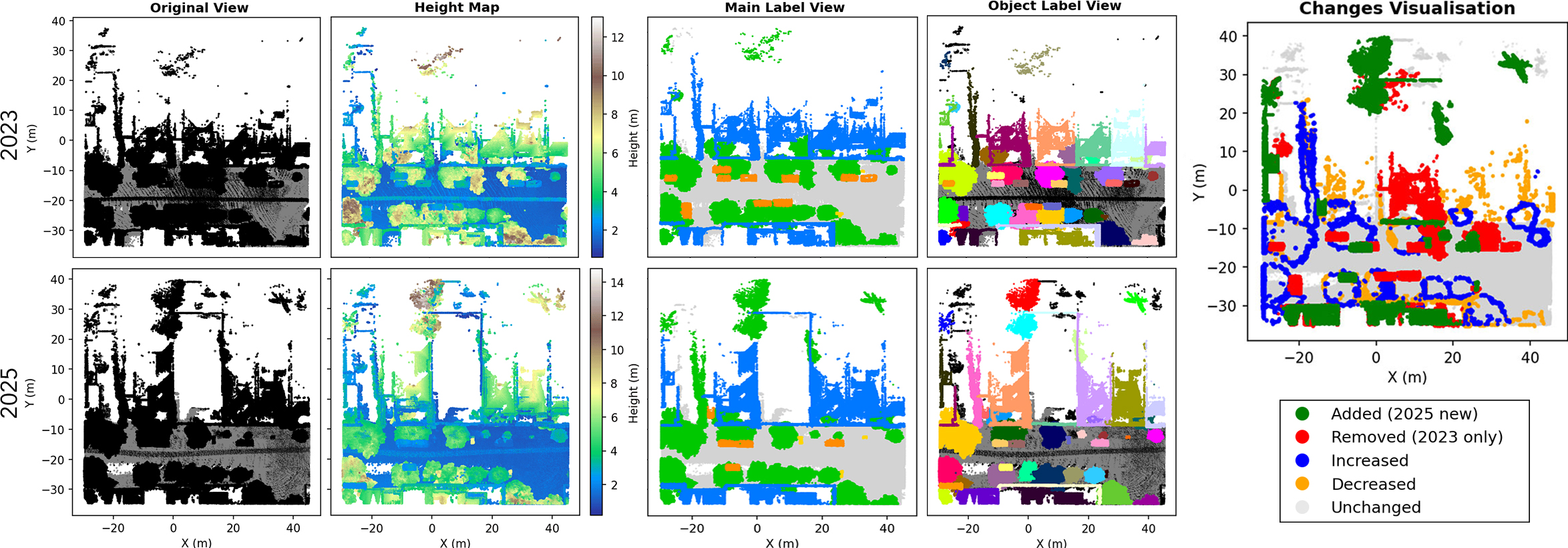}\\[-1pt]
  \includegraphics[width=0.48\textwidth,height=0.12\textheight,keepaspectratio=false]{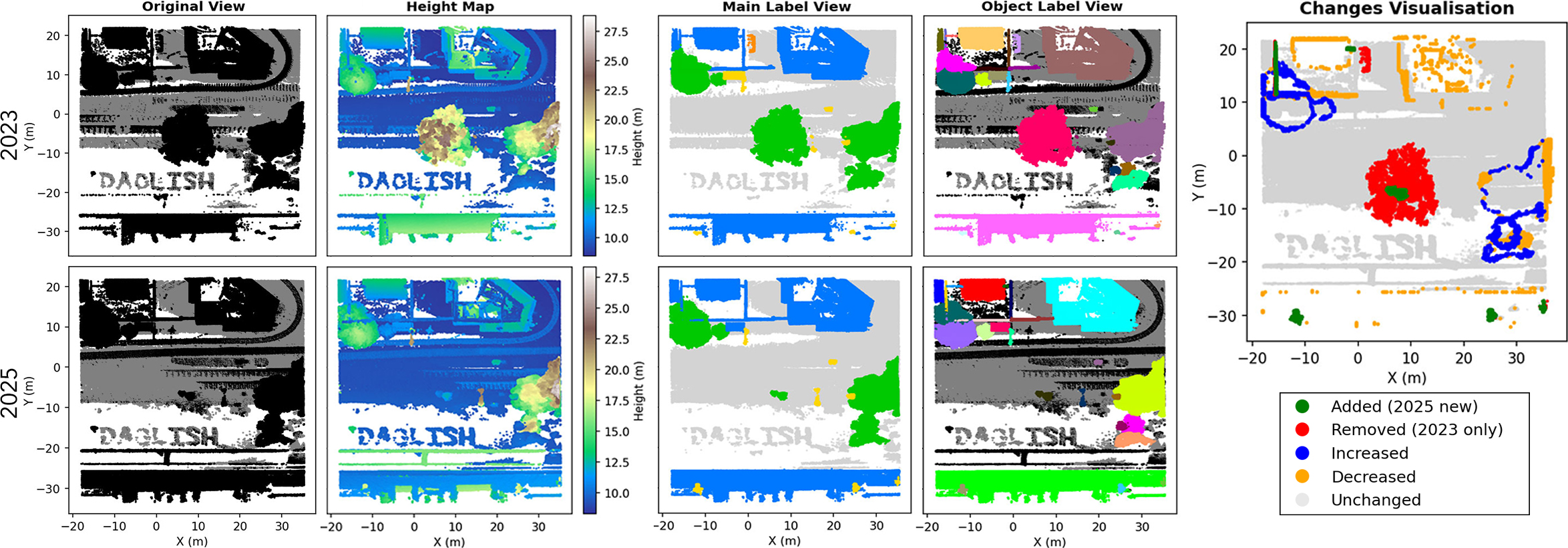}
  \vspace{-6mm}
  \caption{Qualitative results on three Subiaco blocks. For each block, the top and bottom rows show the 2023 and 2025 epochs; columns show the plan view, height map, semantic labels, instances, and object-level change map.}
  \Description{Three paired 2023 and 2025 urban LiDAR examples. Each example contains plan-view point clouds, height maps, semantic labels, instance labels, and a final color-coded change map for added, removed, increased, decreased, and unchanged objects.}
  \label{fig:perth-change-qual}
  \vspace{-5mm}
\end{figure}

\begin{table*}[t]
\centering
\caption{Point-wise change segmentation on Urb3DCD-V2: summary metrics and published class-wise IoU.}
\vspace{-4mm}
\label{tab:urb3dcd_v2}
\footnotesize
\setlength{\tabcolsep}{3.2pt}
\renewcommand{\arraystretch}{1.08}

\begin{tabular*}{\textwidth}{@{\extracolsep{\fill}}
    >{\raggedright\arraybackslash}p{0.46\textwidth}
    >{\raggedright\arraybackslash}p{0.14\textwidth}
    l l c c}
\toprule
\multicolumn{6}{l}{\textbf{Panel A: Summary metrics}} \\
\midrule
Method & Src (Venue, Year) & Rep. & Sup. & mAcc & mIoUch \\
\midrule
DSM-Siamese (FC-Siamese, DSM raster)~\cite{daudt2018fcsn} & IGARSS, 2018 & raster & supervised & 80.91 & 57.41 \\
DSM-FC-EF (FCN early-fusion on DSM)~\cite{daudt2018fcsn} & ICIP, 2018 & raster & supervised & 81.47 & 58.07 \\
SiamKPConv (KPConv Siamese, 3 input feats)~\cite{thomas2019kpconv} & ICCV, 2019 & point & supervised & 91.21 & 80.12 \\
SiamKPConv (KPConv Siamese, 10 feats incl.\ change cues)~\cite{degelis2023needschangeinfo} & CoRR, 2023 & point & supervised & 93.65 & 84.82 \\
SiamKPConv (10 feats, no stabilization)~\cite{degelis2023needschangeinfo} & CoRR, 2023 & point & supervised & 91.44 & 80.49 \\
SiamKPConv (stabilization only, no extra feats)~\cite{degelis2023needschangeinfo} & CoRR, 2023 & point & supervised & 92.92 & 83.80 \\
OneConvFusion (sparse voxel net, one-layer fusion)~\cite{degelis2023needschangeinfo} & CoRR, 2023 & voxel & supervised & 93.71 & 85.19 \\
TripletKPConv (triplet loss, point-wise change)~\cite{degelis2023needschangeinfo} & CoRR, 2023 & point & supervised & 93.67 & 86.34 \\
EncoderFusion SiamKPConv (late fusion of Siam encoders)~\cite{degelis2023needschangeinfo} & CoRR, 2023 & point & supervised & 94.23 & 85.19 \\
EncoderFusion SiamKPConv (10 feats)~\cite{degelis2022siamesekpconv} & ISPRS Ann., 2022 & point & supervised & 94.13 & 85.87 \\
EFSKPConv (KPConv with encoder fusion and selection)~\cite{kharroubi2023dc3dcd} & ISPRS JPRS, 2023 & point & supervised & 94.23 & 85.19 \\
DC3DCD (unsupervised deep clustering on 3D features)~\cite{kharroubi2023dc3dcd} & ISPRS JPRS, 2023 & point & weakly sup. & 68.45 & 57.06 \\
SIREN + S + TVN + TD (object-proxy change cues)~\cite{kharroubi2023dc3dcd} & ISPRS JPRS, 2023 & object proxy & supervised & 95.45 & 82.14 \\
RFF + S + TVN + TD (object-proxy change cues)~\cite{kharroubi2023dc3dcd} & ISPRS JPRS, 2023 & object proxy & supervised & 95.41 & 82.52 \\
PGN3DCD (prior-guided 3DCD network)~\cite{liu2024pgn3dcd} & TGRS, 2024 & point & supervised & 95.92 & 87.05 \\
PointMamba (SSM backbone with Siam head)~\cite{ma2024pointmamba} & NeurIPS, 2024 & point & supervised & 96.02 & 87.62 \\
Mamba3D (SSM 3D encoder with Siam head)~\cite{zhang2024mamba3d} & arXiv, 2024 & point & supervised & 95.85 & 87.20 \\
Serialized Point Mamba (serialized SSM with Siam head)~\cite{wang2024serializedpointmamba} & arXiv, 2024 & point & supervised & 95.98 & 87.47 \\
Ms-DANet (multi-scale difference-aware 3DCD)~\cite{li2025msdanet} & TGRS, 2025 & point & supervised & 96.10 & 87.88 \\
ME-CPT (multi-task enhanced cross-temporal point transformer)~\cite{yang2025mecpt} & TGRS, 2026 & point & supervised & 95.88 & 87.62\\

\textbf{Ours (object-level LoD gating + matching + cue fusion)} & This work & object based & object based & \textbf{96.81} & \textbf{89.52} \\
\bottomrule
\end{tabular*}

\vspace{2mm}

\begin{tabular*}{\textwidth}{@{\extracolsep{\fill}}
    >{\raggedright\arraybackslash}p{0.34\textwidth}
    c c c c c c c}
\toprule
\multicolumn{8}{l}{\textbf{Panel B: Per-class IoU (\%) on published class-wise reports}} \\
\midrule
Method & Unch. & New bld. & Demo. & New veg. & Veg grow. & Miss veg. & Mobile \\
\midrule
SiamKPConv (+10 feat.)~\cite{degelis2023needschangeinfo}  & 97.60 & 92.75 & 86.84 & 95.92 & 68.70 & 73.38 & 94.05 \\
TripletKPConv~\cite{degelis2023needschangeinfo}          & 96.20 & 85.40 & 89.60 & 89.05 & 54.80 & 60.20 & 88.70 \\
EncoderFusion SiamKPConv~\cite{degelis2023needschangeinfo} & 96.46 & 91.00 & 88.42 & 96.31 & 78.82 & 72.68 & 95.16 \\
DC3DCD (EFSKPConv + i.f.)~\cite{kharroubi2023dc3dcd}     & 93.96 & 79.26 & 67.88 & 75.34 & 69.48 & 20.29 & 80.10 \\
\textbf{Ours}                                            & \textbf{98.26} & \textbf{95.61} & \textbf{94.30} & \textbf{95.07} & \textbf{75.87} & \textbf{79.32} & \textbf{95.82} \\
\bottomrule
\end{tabular*}

\vspace{-4mm}
\end{table*}

Each pair receives a label $y\in\mathcal{Y}$. For buildings and vegetation, unmatched instances are \textit{Added} or \textit{Removed}. For matched instances we require class specific overlap and displacement bounds, namely $\mathrm{IoU}_{3\mathrm{D}}>0.12$ and $\|\Delta\mathbf{c}\|_2<1.5$\,m for buildings and $\mathrm{IoU}_{3\mathrm{D}}>0.08$ and $\|\Delta\mathbf{c}\|_2<2.0$\,m for vegetation. A building pair is \textit{Increased} if $\Delta h>0$ and $|\Delta h|>0.60$\,m or if $\Delta V/V^{23}>0.12$, and \textit{Decreased} if $\Delta h<0$ and $|\Delta h|>0.60$\,m or if $\Delta V/V^{23}<-0.12$. It is \textit{Unchanged} if $\mathrm{IoU}_{3\mathrm{D}}>0.20$, $\|\Delta\mathbf{c}\|_2<1.5$\,m, $|\Delta h|\le0.60$\,m and $|\Delta V|/V^{23}\le0.12$, or if all informative statistics stay below the LoD gate. Vegetation uses thresholds $|\Delta h|>0.35$\,m and $|\Delta V|/V^{23}>0.18$ with the same logic. When height and volume indicators disagree we inspect the vertical occupancy histogram inside the overlap and select \textit{Increased} if mass shifts upward and \textit{Decreased} otherwise. Remaining vegetation pairs that satisfy the overlap bound are marked \textit{Unchanged}.

Ground changes are evaluated on a $2$\,m raster. For each tile we compute the median elevation difference $\Delta z$ and the median $\mathrm{LoD}_{95}$. A tile is \textit{Increased} if $\Delta z>0.15$\,m and $|\Delta z|>1.1\,\overline{\mathrm{LoD}}_{95}$ over a connected region larger than $25$\,m$^2$, \textit{Decreased} if $\Delta z<-0.15$\,m under the same conditions, and \textit{Unchanged} otherwise. Ground \textit{Added} and \textit{Removed} are not defined. Mobile objects are treated at the instance level. Unmatched mobiles are labeled \textit{Added} or \textit{Removed}. Matched mobiles are labeled \textit{Unchanged} if $\mathrm{IoU}_{3\mathrm{D}}>0.20$ and $\|\Delta\mathbf{c}\|_2<2.0$\,m. Otherwise, the pair is assigned to the epoch with the larger unmatched occupancy inside the union box, yielding \textit{Added} when the 2025 support dominates and \textit{Removed} otherwise. No annotated mobile instance is discarded during evaluation.

\begin{table*}[t]
\centering
\caption{Ablation on LoDA: impact of LoD-aware registration, correspondence, instance formation, and change cues. Metrics: ACC, mF1, mIoU, and per-class IoU (\%).}
\vspace{-2mm}
\label{tab:ablation}
\footnotesize
\setlength{\tabcolsep}{3.4pt}
\renewcommand{\arraystretch}{1.05}

\begin{tabularx}{\textwidth}{@{} l >{\raggedright\arraybackslash}X c c c c c c c c @{}}
\toprule
Category & Variant & ACC & mF1 & mIoU & Added & Removed & Increased & Decreased & Unchanged \\
\midrule
Baseline (learning) &
EFS KPConv & 94.6 & 86.4 & 74.3 & 88.3 & 76.8 & 54.1 & 58.4 & 93.8 \\
\midrule
Reference (ours) &
Full pipeline (LoD-aware registration + proxies + cut-pursuit instances + $\Delta h$,$\Delta V$,$D_{\perp}$) & 95.0 & 90.8 & 83.0 & 87.8 & 81.2 & 77.2 & 73.7 & 95.1 \\
\midrule
\multirow{5}{*}{\parbox{1.55cm}{\centering Registration \\ and LoD}} &
No LoD gating in decisions & 93.6 & 86.0 & 75.2 & 86.0 & 78.5 & 60.4 & 57.8 & 93.2 \\
& Global LoD (single scalar, no spatial variation) & 94.3 & 89.0 & 80.6 & 87.0 & 80.0 & 72.4 & 68.9 & 94.5 \\
& LoD without pose covariance term ($n^\top\Sigma_t n$ removed) & 94.5 & 89.6 & 81.7 & 87.3 & 80.6 & 74.9 & 71.0 & 94.8 \\
& ICP without robust loss (L2 point-to-plane) & 94.2 & 89.2 & 81.0 & 86.8 & 79.6 & 74.1 & 70.2 & 94.3 \\
& No vertical alignment and height normalisation & 93.8 & 87.2 & 77.5 & 86.1 & 79.0 & 65.5 & 62.8 & 93.9 \\
\midrule
\multirow{2}{*}{\parbox{1.55cm}{\centering Correspondence}} &
No geometry-only proxies (match final instances directly) & 94.0 & 88.5 & 79.9 & 84.5 & 76.0 & 75.0 & 70.4 & 93.8 \\
& Greedy matching (no Hungarian assignment) & 94.8 & 90.1 & 82.3 & 87.4 & 80.4 & 76.2 & 72.5 & 95.0 \\
\midrule
\multirow{3}{*}{\parbox{1.55cm}{\centering Instance \\ formation}} &
No cut-pursuit superpoints (voxel connected components only) & 94.1 & 88.8 & 80.0 & 85.0 & 77.5 & 73.8 & 69.9 & 94.0 \\
& No planar merging for buildings (patch instances kept separate) & 94.4 & 89.1 & 80.9 & 86.2 & 78.8 & 74.5 & 70.7 & 94.4 \\
& Fixed-radius vegetation clustering (no density adaptation) & 94.7 & 89.9 & 82.1 & 87.1 & 80.7 & 75.9 & 72.1 & 94.7 \\
\midrule
\multirow{4}{*}{\parbox{1.55cm}{\centering Change \\ cues}} &
Without height cue $\Delta h$ & 94.1 & 88.1 & 79.2 & 87.2 & 80.9 & 68.3 & 64.9 & 94.9 \\
& Without volume cue $\Delta V$ & 94.2 & 88.6 & 79.9 & 87.0 & 80.1 & 70.5 & 67.1 & 94.7 \\
& Without normal-direction cue $D_{\perp}$ & 94.6 & 89.4 & 81.2 & 87.6 & 81.0 & 73.1 & 69.2 & 94.9 \\
& Height-only cue ($\Delta h$ only, no $\Delta V$ and no $D_{\perp}$) & 93.7 & 86.8 & 77.0 & 86.8 & 79.7 & 63.9 & 60.5 & 94.2 \\
\bottomrule
\end{tabularx}
\vspace{-4mm}
\end{table*}

\begin{table}[t]
\centering
\caption{Efficiency profile on the 15 LoDA evaluation blocks. Time is wall-clock seconds. Breakdown reports registration, segmentation, matching, and change analysis as R/S/M/C. PeakMem is peak host RAM (GB). Pts is the fused point count per block (million). Inst is the number of non-ground instances used for matching and change analysis.}
\vspace{-2mm}
\label{tab:efficiency_loda}
\scriptsize
\setlength{\tabcolsep}{3.2pt}
\renewcommand{\arraystretch}{1.05}
\resizebox{\columnwidth}{!}{%
\begin{tabular}{lrrrrrr}
\toprule
Block & Pts (M) & Inst & Breakdown (R/S/M/C) & Total (s) & PeakMem (GB) \\
\midrule
C01 & 1.12 & 240 & 10.2/27.5/1.9/8.0  & 47.6 & 9.8 \\
C02 & 1.28 & 275 & 10.8/30.6/2.1/8.6  & 52.1 & 10.7 \\
C03 & 1.45 & 310 & 11.7/33.2/2.3/9.2  & 56.4 & 11.6 \\
C04 & 1.62 & 345 & 12.5/35.7/2.5/10.1 & 60.8 & 12.5 \\
C05 & 1.38 & 290 & 11.4/32.1/2.2/9.0  & 54.7 & 11.2 \\
C06 & 1.95 & 420 & 14.0/41.5/3.0/11.8 & 70.3 & 15.4 \\
C07 & 1.74 & 380 & 13.1/38.8/2.8/11.0 & 65.7 & 14.2 \\
C08 & 1.56 & 330 & 12.2/35.0/2.5/9.8  & 59.5 & 12.7 \\
C09 & 2.18 & 460 & 14.9/45.1/3.2/12.6 & 75.8 & 16.8 \\
C10 & 1.33 & 285 & 11.0/31.2/2.1/8.7  & 53.0 & 10.9 \\
C11 & 2.05 & 440 & 14.4/42.9/3.1/12.0 & 72.4 & 16.1 \\
C12 & 1.87 & 405 & 13.6/40.2/2.9/11.4 & 68.1 & 15.0 \\
C13 & 1.22 & 255 & 10.6/29.0/2.0/8.3  & 49.9 & 10.3 \\
C14 & 1.69 & 365 & 12.9/37.6/2.7/10.7 & 63.9 & 13.8 \\
C15 & 1.51 & 320 & 12.0/34.1/2.4/9.5  & 58.0 & 12.3 \\
\midrule
\textbf{Avg.} & \textbf{1.60} & \textbf{341} & \textbf{12.4/35.6/2.5/10.0} & \textbf{60.5} & \textbf{12.9} \\
\bottomrule
\end{tabular}%
}
\vspace{-6mm}
\end{table}

For each matched pair, we compute a deterministic confidence score in $[0,1]$ from LoD-normalized height, volume, overlap, centroid-shift, and normal-displacement cues. Class-specific weights are selected once on the LoDA validation split and then frozen for all experiments. The score is used only for ranking and manual inspection and is not interpreted as a calibrated posterior probability.

\vspace{-2mm}
\subsection{Evaluation protocol}
\label{sec:Evaluation-protocol}

Metrics are computed after aggregating predictions over the full test split. The evaluation set contains every annotated non-ground instance, including all mobile instances. Ground tiles are excluded from Tables~\ref{tab:method_comp_perth} and~\ref{tab:perth_blocks}. Let $TP_y$, $FP_y$, and $FN_y$ denote the counts for label $y\in\mathcal{Y}$. We report micro accuracy as $\mathrm{ACC}=\sum_{y\in\mathcal{Y}} TP_y / N$, where $N$ is the number of evaluated instances. Per-class scores are $\mathrm{IoU}_y = TP_y / (TP_y + FP_y + FN_y)$ and $\mathrm{F1}_y = 2TP_y / (2TP_y + FP_y + FN_y)$. Macro IoU and macro F1 are the unweighted mean over the five labels in $\mathcal{Y}$ and are not frequency-weighted. If a method emits no valid prediction for an annotated instance, that instance is counted as a false negative for its ground-truth label.

\section{Experiments}\label{sec:experiments}
We assess the pipeline on the LoDA benchmark by running end-to-end object-level change inference on the fused multi-temporal maps and reporting label-level accuracy, macro F1, and macro IoU against representative raster, point, voxel, and deep-learning baselines.

\subsection{Experimental Setup}
All experiments were conducted on a single Linux workstation running Ubuntu~22.04, equipped with an Intel Xeon Gold~6330 CPU (28 cores), 64~GB RAM, and an NVIDIA GeForce RTX~5090 GPU with 24~GB VRAM. We used a Conda environment with Python~3.10, PyTorch~2.1.2, CUDA~12.1, and cuDNN~9.0. Point cloud preprocessing, registration, and I/O were implemented with Open3D~\cite{zhou2018open3d}~0.18, PCL~\cite{rusu2011pcl}~1.13, and PDAL~2.6. Learning-based baselines were trained from their official public codebases (MinkowskiEngine~0.5.4 for sparse voxels and the released KPConv implementation for point-based networks).

All methods are evaluated on the same $120\,\mathrm{m}\times120\,\mathrm{m}$ blocks and the same train/val/test split, using identical shared preprocessing: Statistical Outlier Removal, rigid alignment with $T_{23\rightarrow25}$, and ground removal + height normalization via the $2\,\mathrm{m}$ ground model. Our spatially varying $\mathrm{LoD}_{95}$ gating is \emph{not} applied to any baseline. All rule thresholds in the proposed pipeline, including semantic criteria, clustering limits, and change thresholds, were selected once on the LoDA validation split and then frozen for LoDA test and Urb3DCD-V2, with no per-scene or per-dataset retuning. Raster DSM baselines use $0.5\,\mathrm{m}$ nDSM grids ($240\times240$ per block), sparse-voxel baselines voxelize at $0.25\,\mathrm{m}$, and point-based baselines use $0.25\,\mathrm{m}$ downsampled points with uniform $N{=}8192$ sampling per epoch-block when needed. Unless specified by the original codebase, learning baselines use AdamW~\cite{loshchilov2019adamw} (lr $10^{-3}$, weight decay $10^{-4}$) with cosine decay~\cite{loshchilov2017sgdr} for 100 epochs (batch size 2), standard geometric augmentation (yaw rotation, scaling $[0.95,1.05]$, jitter $\sigma{=}0.01\,\mathrm{m}$, dropout), and early stopping on validation mIoU (patience 20); results are averaged over three seeds, while our pipeline is deterministic. For Urb3DCD-V2, we follow the official split and point-wise evaluation protocol.

\subsection{Evaluation on LoDA}
\label{sec:baseline-comp}
Tables~\ref{tab:method_comp_perth} and~\ref{tab:perth_blocks} summarise the LoDA results. Our method achieves $95.0\%$ ACC, $90.8\%$ mF1, and $83.0\%$ mIoU. Compared with the strongest baseline, EFS KPConv, mIoU and mF1 improve by $8.7$ and $4.4$ points. The largest gains occur for \textit{Increased} and \textit{Decreased}, with improvements of $23.1$ and $15.3$ IoU points. Fig.~\ref{fig:perth-change-qual} shows qualitative results. Typical failures involve sparse tree crowns, fragmented facade patches, and short-lived parked vehicles. Because compatible five-label object-level baselines are unavailable, Table~\ref{tab:urb3dcd_v2} also reports the closest published object-proxy comparisons on Urb3DCD-V2.

\subsection{Efficiency and Scalability}
\label{sec:efficiency}

Table~\ref{tab:efficiency_loda} reports the runtime and memory usage. Across 15 LoDA blocks, pipeline processes an average of $1.60$ million points and $341$ non-ground instances per block in $60.5\,\mathrm{s}$, with $12.9\,\mathrm{GB}$ peak host memory. Segmentation is the dominant cost, requiring $35.6\,\mathrm{s}$ and accounting for $58.8\%$ of the average runtime, whereas object matching requires only $2.5\,\mathrm{s}$ ($4.1\%$). From C01 to C09, the input increases from $1.12$ to $2.18$ million points, while runtime and memory increase from $47.6$ to $75.8\,\mathrm{s}$ and from $9.8$ to $16.8\,\mathrm{GB}$, respectively, indicating consistent scaling over the evaluated block range.

\subsection{Evaluation on Urb3DCD-V2}

We further evaluate on Urb3DCD-V2 under the official point-wise protocol by projecting each instance-level decision to point labels. Table~\ref{tab:urb3dcd_v2} reports $96.81\%$ mAcc and $89.52\%$ mIoUch, exceeding the best listed baselines by $1.36$ and $3.18$ points, respectively.


\subsection{Ablation Study}

Table~\ref{tab:ablation} isolates the effects of LoD-aware registration, correspondence, instance formation, and change cues. The full pipeline achieves $83.0$ mIoU and $90.8$ mF1. Removing LoD gating causes the largest mIoU drop, mainly increasing confusion between \textit{Increased} and \textit{Decreased}.

\vspace{-2mm}
\section{Conclusion}
\label{sec:conclusion}
We presented LoDA and an LoD-aware object-level change detection pipeline for bi-temporal vehicle LiDAR maps. The method achieves $95.0\%$ ACC, $90.8\%$ mF1, and $83.0\%$ mIoU on LoDA, and $96.81\%$ mAcc and $89.52\%$ mIoUch on Urb3DCD-V2, showing that LoD-gated instance-level reasoning improves reliable map update under density variation and residual misregistration.

\begin{acks}
This work was supported by Western Australia Machine Intelligence Group Pty Ltd under the project ``Building, Correcting, Planning and Updating High-Fidelity 3D City Maps for Western Australia'' (WG-20267524-CM). The project focuses on the planning, construction, correction, and long-term updating of high-fidelity 3D urban and road maps. In particular, it investigates the generation of planned or future road environments from infrastructure development plans and existing geospatial information, including LiDAR point clouds, GNSS data, and previously constructed 3D maps. This support facilitated the development and evaluation of the LoDA benchmark and the proposed object-level change-detection framework. The authors gratefully acknowledge this support.
\end{acks}

\vspace{-2mm}
\bibliographystyle{ACM-Reference-Format}
\bibliography{main}

\end{document}